\pdfoutput=1
\documentclass[11pt]{article}

\usepackage[]{acl}

\usepackage{times}
\usepackage{latexsym}

\usepackage[T1]{fontenc}
\usepackage[utf8]{inputenc}

\usepackage{microtype}
\usepackage{times}
\usepackage{latexsym}
\usepackage{algorithm}
\usepackage{algorithmic}
\usepackage[switch]{lineno} 
\usepackage[utf8]{inputenc}
\usepackage{CJKutf8}
\usepackage{multirow}
\usepackage{booktabs}
\usepackage{graphicx}
\usepackage{amsmath}
\usepackage{amssymb}
\usepackage{xpinyin}
\usepackage{pifont}
\usepackage{subfigure}
\usepackage{pdfpages}
\usepackage{CJK}
\newcommand{\MethodName}{CL}

\usepackage{xcolor}

\title{Contextual Similarity is More Valuable than Character Similarity: \\ An Empirical Study for Chinese Spell Checking}

\author{Ding Zhang$^{1}$\thanks{ $^*$ indicates equal contribution.},~~Yinghui Li$^{1*}$,~Qingyu Zhou$^{2}$,~Shirong Ma$^{1}$,\\~\textbf{Yangning Li}$^{1,4}$, ~\textbf{Yunbo Cao}$^{3}$ \and \textbf{Hai-Tao Zheng}$^{1,4}$\thanks{ $^{\dagger}$ Corresponding author: Hai-Tao Zheng. (E-mail: zheng.haitao@sz.tsinghua.edu.cn)} \\
        $^{1}$Tsinghua Shenzhen International Graduate School, Tsinghua University \\ 
        $^{2}$OPPO Research Institute, $^{3}$Tencent Cloud Xiaowei, $^{4}$Peng Cheng Laboratory \\
        \texttt{zhangdingthu@outlook.com}, \texttt{liyinghu20@mails.tsinghua.edu.cn}
        }

\begin{document}
\maketitle
\begin{abstract}
Chinese Spell Checking (CSC) task aims to detect and correct Chinese spelling errors. Recently, related researches focus on introducing character similarity from confusion set to enhance the CSC models, ignoring the context of characters that contain richer information. To make better use of contextual information, we propose a simple yet effective Curriculum Learning (CL) framework for the CSC task. With the help of our model-agnostic CL framework, existing CSC models will be trained from easy to difficult as humans learn Chinese characters and achieve further performance improvements. Extensive experiments and detailed analyses on widely used SIGHAN datasets show that our method outperforms previous state-of-the-art methods. More instructively, our study empirically suggests that contextual similarity is more valuable than character similarity for the CSC task.
\end{abstract}
%
% \begin{keywords}
% Natural Language Processing, Chinese Spell Checking,  Contextual Similarity, Curriculum Learning
% \end{keywords}
%
\section{Introduction}
\label{sec:intro}

\begin{CJK*}{UTF8}{gbsn}
Chinese Spell Checking (CSC) aims to detect and correct spelling errors contained in Chinese text~\cite{DBLP:conf/acl/LiZLLLSWLCZ22, ma-etal-2022-linguistic}. CSC is receiving more and more attention because it benefits many applications, such as essay scoring~\cite{DBLP:conf/emnlp/DongZ16}, OCR~\cite{DBLP:conf/lrec/AfliQWS16}, and ASR~\cite{DBLP:conf/emnlp/WangSLHZ18}. As a fundamental NLP task, CSC is challenging because the Chinese spelling errors are mainly caused by confusing characters, i.e., phonologically/visually similar characters~\cite{DBLP:conf/acl/LiuYYZW20}. As shown in Table~\ref{tab:intro}, “曰(\textit{\pinyin{yue1}, say)}” and “日(\textit{\pinyin{ri4}, day)}” are confusing due to their similar strokes.
\begin{CJK*}{UTF8}{gbsn}

\end{CJK*}

To enable CSC models to handle confusion characters better, the pre-defined confusion set (i.e., the set of phonologically/visually similar characters) has been long regarded as a good external resource. Many previous works~\cite{DBLP:conf/acl/LiuYYZW20,DBLP:conf/acl/ZhangHLL20} have aimed to leverage the confusion set to introduce phonological/visual similarities into the CSC models. 
\emph{However, these existing methods simply focus on the character similarity provided by the confusion set, but ignore the context of the characters.} 
In fact, in a sentence with a spelling error, the context of the error position provides more useful information that facilitates the CSC process.
For example, in Table~\ref{tab:intro}, if the model pays attention to “帽子 (hat)” in the context, it will easily associate the wrong character “带(\textit{\pinyin{dai4}}, bring)” with the correct character “戴(\textit{\pinyin{dai4}}, wear)”.
\emph{Therefore, we believe that the contextual similarity of the characters is more important for CSC than the character similarity.}

\end{CJK*}

\begin{CJK*}{UTF8}{gbsn}

\begin{table}[t]
\centering
\small
\renewcommand\arraystretch{1.2}
\begin{tabular}{|c|c|}% 通过添加 | 来表示是否需要绘制竖线
\hline  % 在表格最上方绘制横线
Error Type & Phonetically Similar  (83\%)\\
\hline  % 在表格最上方绘制横线
Input & 他\textcolor{red}{带(\textit{\pinyin{dai4}, bring)}}着一顶帽子。\\
\hline 
Correct& 他\textcolor{blue}{戴(\textit{\pinyin{dai4}, wear})}着一顶帽子。\\
\hline
Translation & He \textcolor{blue}{\underline{\textit{wears}}} a hat.\\
\hline
% \hline
% \\

\hline  % 在表格最上方绘制横线
Error Type&Visually Similar  (48\%)\\
\hline  % 在表格最上方绘制横线
Input & 
今天是个好\textcolor{red}{曰(\textit{\pinyin{yue1}, say)}}子。\\
\hline 
Correct& 今天是个好\textcolor{blue}{日(\textit{\pinyin{ri4}, day})}子。\\
\hline
Translation & Today is a good
 \textcolor{blue}{\underline{\textit{day}}}.
\\
\hline
\end{tabular}

\caption{Examples of Chinese spelling errors. The \textcolor{red}{wrong}/\textcolor{blue}{correct} characters are in \textcolor{red}{red}/\textcolor{blue}{blue}.}
\label{tab:intro}
\end{table}
\end{CJK*}

\begin{figure*}[]
\centering
% \includepdf[scale=1.3,delta=0mm 5mm,height=0.48\textwidth]{model.pdf}
% width=\textwidth
\includegraphics[width=0.90\textwidth]{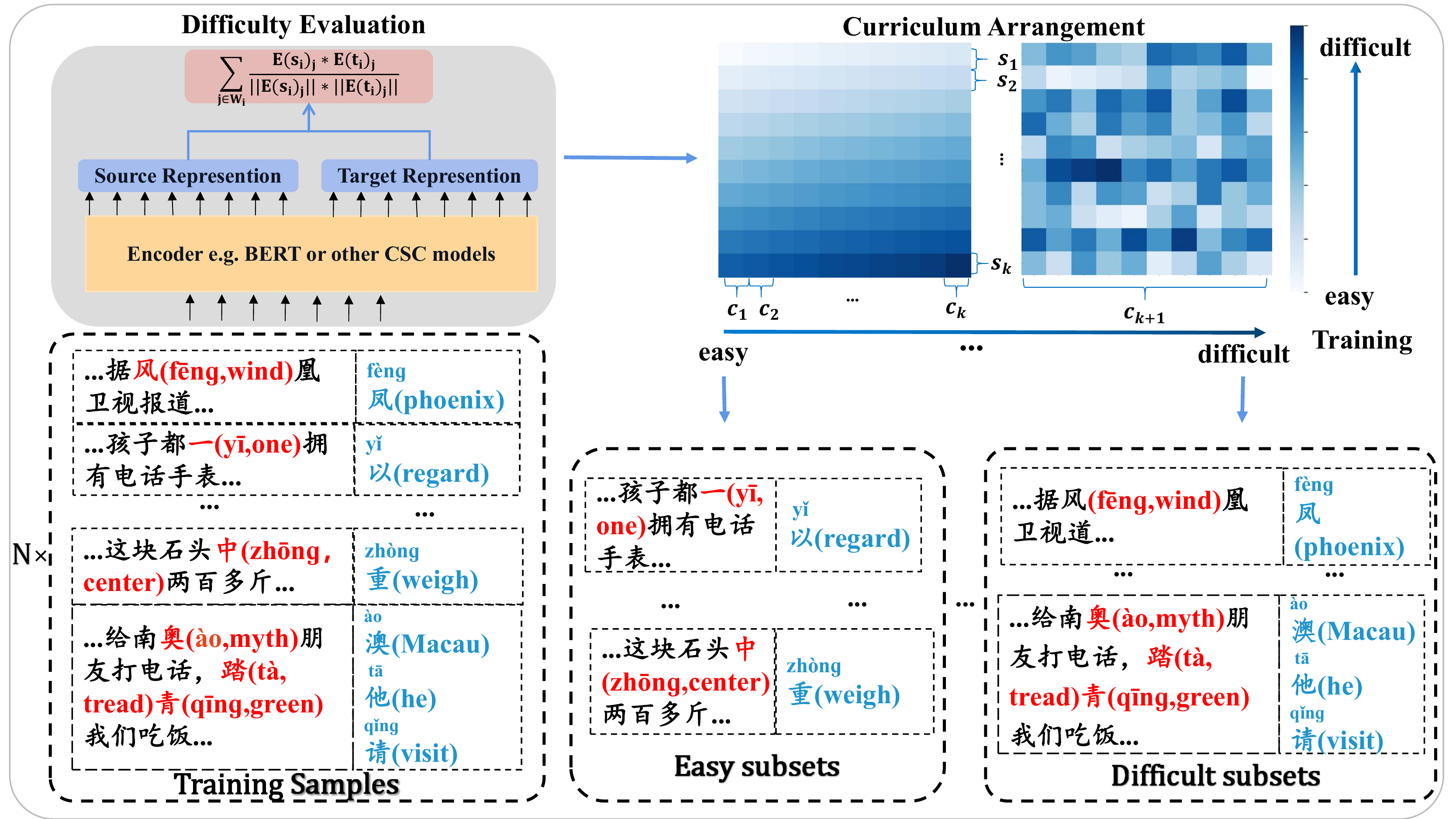}

\caption{Overview of our proposed curriculum learning (CL) framework for the CSC task.}
\label{Method_Figure}
\end{figure*}

In this paper, we aim to enhance CSC models by introducing contextual similarity of Chinese characters. 
Considering that the CSC task itself is inseparable from human learning, we hope that the model can learn like a human learns to correct spelling errors.
We all know that for a student who is just beginning to learn Chinese characters, the teacher always teaches him or her from easy to difficult. 
Therefore, inspired by the process of humans learning Chinese characters, we also want to guide the model to learn from easy to hard. And this motivation just coincides with curriculum learning.

The core idea of curriculum learning is to train models from easy to hard~\cite{soviany2021curriculum}. And the key to curriculum learning is to design a mechanism to measure the difficulty of samples. Benefiting from this mechanism, we naturally use the contextual similarity of characters as the metric for measuring the sample's difficulty, so as to organize the scattered training samples into ordered samples for model training. Specifically, we train the model in the order from samples with low contextual similarity to samples with high contextual similarity. Hence, the model achieves better performance than only using the traditional character similarity of confusion set.
Moreover, our curriculum learning framework is model-agnostic so that it brings stable improvements for most existing CSC models.

The contributions of our work are summarized as: 
(1) We empirically verify that contextual similarity is more valuable than character similarity in the CSC task, which is instructive for future works. 
(2) We propose a simple yet effective curriculum learning framework that enhances the CSC models to explicitly focus on the contextual similarity between Chinese characters. 
(3) We achieve new state-of-the-art performance on SIGHAN benchmarks and conduct extensive analyses to demonstrate the effectiveness of our proposed method.

\begin{table*}[h]
\small
\centering
\begin{tabular}{@{}c|l|p{0.65cm}p{0.65cm}p{0.65cm}p{0.65cm}|p{0.65cm}p{0.65cm}p{0.65cm}p{0.65cm}@{}}
\toprule
\multirow{2}{*}{Dataset} & \multicolumn{1}{c|}{\multirow{2}{*}{Method}} & \multicolumn{4}{c|}{Detection Level} & \multicolumn{4}{c}{Correction Level} \\
 & \multicolumn{1}{c|}{}  & Acc & Pre & Rec & F1 & Acc  & Pre & Rec & F1 \\ \midrule
 
\multicolumn{1}{l|}{\multirow{9}{*}{SIGHAN13}} & 
SpellGCN~\cite{DBLP:conf/acl/ChengXCJWWCQ20} &-& 80.1 & 74.4 & 77.2 &-&  78.3 & 72.7 & 75.4 \\
\multicolumn{1}{l|}{} & MLM-phonetics~\cite{DBLP:conf/acl/ZhangPZWHSWW21} &-&  82.0 & 78.3 & 80.1 &-&  79.5 & 77.0 & 78.2 \\ 
\multicolumn{1}{l|}{} &REALISE~\cite{DBLP:conf/acl/XuLZLWCHM21} &\underline{\textbf{82.7}}&  {88.6} & \textbf{\underline{82.5}} & \underline{85.4} &\underline{\textbf{81.4}}&  {87.2} & \underline{\textbf{81.2}} & \underline{84.1} \\
\cmidrule(l){2-10} 
\multicolumn{1}{l|}{} & Soft-Masked BERT~\cite{DBLP:conf/acl/ZhangHLL20} &-&  81.1 & 75.7 & 78.3&- &  75.1 & 70.1 & 72.5 \\
\multicolumn{1}{l|}{} & \MethodName{}~(Soft-Masked BERT) &-&  84.7$^\uparrow$ & 77.0$^\uparrow$ & 80.7$^\uparrow$ &-&  80.9$^\uparrow$ & 74.5$^\uparrow$ & 77.6$^\uparrow$\\
\cmidrule(l){2-10} 
\multicolumn{1}{l|}{} & BERT~\cite{DBLP:conf/acl/HuangLJZCWX20}  &70.6& 98.7& 70.6 & 82.3  &67.8& 98.6 & 67.8 & 80.4 \\

\multicolumn{1}{l|}{} & \MethodName{}~(BERT)&75.4$^\uparrow$& 99.1$^\uparrow$ & 74.8$^\uparrow$ & 85.3$^\uparrow$ &74.9$^\uparrow$&  99.1$^\uparrow$ & 73.2$^\uparrow$ & 84.2$^\uparrow$ \\ 
\cmidrule(l){2-10} 
\multicolumn{1}{l|}{} & MacBERT~\cite{DBLP:conf/emnlp/CuiC000H20}&70.8 &  \underline{98.7} & 70.8 & 82.5 & 68.0 &  \underline{98.6} & 67.9 & 80.4 \\
\multicolumn{1}{l|}{} & \MethodName{}~(MacBERT)&76.3$^\uparrow$&  \textbf{99.3}$^\uparrow$ & 75.7$^\uparrow$ & \textbf{85.9}$^\uparrow$&75.8$^\uparrow$ & \textbf{99.2}$^\uparrow$ & 73.8$^\uparrow$ & \textbf{84.6}$^\uparrow$ \\

\midrule
\multicolumn{1}{l|}{\multirow{9}{*}{SIGHAN14}} & SpellGCN~\cite{DBLP:conf/acl/ChengXCJWWCQ20}  &-& 65.1 & 69.5 & 67.2  &-& 63.1 & 67.2 & 65.3 \\
\multicolumn{1}{l|}{} &  MLM-phonetics~\cite{DBLP:conf/acl/ZhangPZWHSWW21} &-& 66.2 & \underline{73.8} & \underline{69.8}  &-& 64.2 & \textbf{\underline{73.8}} & \underline{68.7} \\
\multicolumn{1}{l|}{} & REALISE~\cite{DBLP:conf/acl/XuLZLWCHM21} &\underline{\textbf{78.4}}& {67.8} & {71.5} & {69.6} &\underline{\textbf{77.7}}&  {66.3} & {70.0} & {68.1} \\
\cmidrule(l){2-10} 
\multicolumn{1}{l|}{} & Soft-Masked BERT~\cite{DBLP:conf/acl/ZhangHLL20}  &-& 65.2 & 70.4 & 67.7  &-& 63.7 & 68.7 & 66.1 \\
\multicolumn{1}{l|}{} & \MethodName{}~(Soft-Masked BERT) &-& 68.4$^\uparrow$ & 70.9$^\uparrow$ & 69.6$^\uparrow$ &-&  67.8$^\uparrow$ &69.1$^\uparrow$ & 68.4$^\uparrow$\\
\cmidrule(l){2-10} 
\multicolumn{1}{l|}{} & BERT~\cite{DBLP:conf/acl/HuangLJZCWX20}  &72.7& {78.6} & 60.7 & 68.5  &71.2& 77.8 & 57.6 & 66.2 \\
\multicolumn{1}{l|}{} & \MethodName{}~(BERT) &75.8$^\uparrow$ &  79.2$^\uparrow$ & 61.6$^\uparrow$ & 69.3$^\uparrow$ &74.8$^\uparrow$ &  78.5$^\uparrow$ & 60.8$^\uparrow$ & 68.5$^\uparrow$ \\ 
\cmidrule(l){2-10} 
\multicolumn{1}{l|}{} & MacBERT~\cite{DBLP:conf/emnlp/CuiC000H20} &72.9& \underline{78.8} & {61.0} & {68.8} &71.5&  \underline{78.0} & {58.0} & {66.5} \\
\multicolumn{1}{l|}{} & \MethodName{}~(MacBERT) &76.1$^\uparrow$ &  \textbf{79.7}$^\uparrow$ & 62.4$^\uparrow$ & \textbf{70.0}$^\uparrow$ &{75.0}$^\uparrow$ &  \textbf{79.0}$^\uparrow$ & 61.4$^\uparrow$ & \textbf{69.1}$^\uparrow$ \\ 

\bottomrule
 
\multicolumn{1}{l|}{\multirow{10}{*}{SIGHAN15}} &SpellGCN~\cite{DBLP:conf/acl/ChengXCJWWCQ20}  &-& 74.8 & 80.7 & 77.7  &-& 72.1 & 77.7 & 75.9 \\
\multicolumn{1}{l|}{} & PLOME~\cite{DBLP:conf/acl/LiuYYZW20}&-&
77.4& 81.5 &79.4 &-&75.3 &79.3& 77.2\\
\multicolumn{1}{l|}{} & MLM-phonetics~\cite{DBLP:conf/acl/ZhangPZWHSWW21}  &-& {77.5} & \underline{\textbf{83.1}} & \underline{80.2}  &-& {74.9} & \underline{\textbf{80.2}} & 77.5 \\ 
\multicolumn{1}{l|}{} & REALISE~\cite{DBLP:conf/acl/XuLZLWCHM21}  &\textbf{\underline{84.7}}& 77.3 & 81.3 & 79.3  &\underline{\textbf{84.0}}& {75.9} & {79.9} & \underline{77.8} \\
\cmidrule(l){2-10} 
\multicolumn{1}{l|}{} & Soft-Masked BERT~\cite{DBLP:conf/acl/ZhangHLL20}  &-& 73.7 & 73.2 & 73.5  &-& 66.7 & 66.2 & 66.4 \\
\multicolumn{1}{l|}{} & \MethodName{}~(Soft-Masked BERT) &-& 83.5$^\uparrow$ & 74.8$^\uparrow$ & 78.9$^\uparrow$ &-&  79.9$^\uparrow$ & 72.1$^\uparrow$ & 75.8$^\uparrow$ \\
\cmidrule(l){2-10} 
\multicolumn{1}{l|}{} & BERT~\cite{DBLP:conf/acl/HuangLJZCWX20}  &79.9& 84.1 & 72.9 & 78.1  &77.5& 83.1 & 68.0 & 74.8 \\
\multicolumn{1}{l|}{} & \MethodName{}~(BERT) &80.5$^\uparrow$& {85.0}$^\uparrow$ & 74.5$^\uparrow$ & {79.4}$^\uparrow$ &79.0$^\uparrow$&  {84.2}$^\uparrow$ &72.3$^\uparrow$ & 77.8$^\uparrow$ \\
\cmidrule(l){2-10} 
\multicolumn{1}{l|}{} & MacBERT~\cite{DBLP:conf/emnlp/CuiC000H20}  &80.0& \underline{84.3} & 73.1 & 78.3 &77.7& \underline{83.3} & {68.2} & 75.0 \\
\multicolumn{1}{l|}{} & \MethodName{}~(MacBERT)&80.9$^\uparrow$&  \textbf{85.8}$^\uparrow$ & 75.4$^\uparrow$ &  \textbf{80.3}$^\uparrow$ &79.3$^\uparrow$& \textbf{84.7}$^\uparrow$ & 73.0$^\uparrow$ & \textbf{78.4}$^\uparrow$ \\
\bottomrule
\end{tabular}

\caption{The performance of our CL method and all baselines. "$^\uparrow$" indicates that our CL method is able to enhance the corresponding baseline. We \underline{underline} the previous state-of-the-art performance for convenience.}
\label{Main_Results}
\end{table*}

\section{Related Work}
\label{sec2}

\subsection{Chinese Spell Checking}

In the field of CSC~\cite{li-etal-2022-learning-dictionary}, many works focus on constructing and
employing confusion set to guide the models to correct the erroneous characters.
%Pointer Network
% Pointer Networks~\cite{DBLP:conf/acl/WangTZ19} incorporates a copy mechanism into a Seq2Seq model and infuses the network with off-the-shelf confusion set~\cite{DBLP:conf/emnlp/WangSLHZ18} for acquiring corrected characters.
%SpellGCN
SpellGCN~\cite{DBLP:conf/acl/ChengXCJWWCQ20} employs graph convolutional network and pre-defined confusion set~\cite{DBLP:conf/acl-sighan/WuLL13} to generate candidate characters for the CSC task. 
PLOME~\cite{DBLP:conf/acl/LiuYYZW20} and MLM-phonetics~\cite{DBLP:conf/acl/ZhangPZWHSWW21} optimize the masking mechanism of masked language models via confusion set and achieves previous state-of-the-art performance.
Besides, REALISE~\cite{DBLP:conf/acl/XuLZLWCHM21} designs BERT-based fusion network to capture multi-modal information, including phonetic and graphic knowledge.

To the best of our knowledge, existing CSC works improve model performance by introducing character
similarity provided by confusion set, but has not made the model focus on contextual similarity of Chinese characters. As a matter of fact, the context of the spelling error position is able to provide vital information for CSC task. In this paper, it is the first time that contextual similarity is applied successfully into the CSC task.

\subsection{Curriculum Learning}
The idea of Curriculum Learning is to train models from easy to hard, which is proposed in~\cite{DBLP:conf/icml/BengioLCW09}. With the great success in CV~\cite{DBLP:journals/corr/abs-2302-09328, DBLP:journals/corr/abs-2301-05997}, Curriculum Learning has attracted many researchers to apply this strategy to all kinds of NLP~\cite{DBLP:journals/corr/abs-2211-04023} tasks, which include Machine Translation~\cite{kocmi2017curriculum, DBLP:journals/corr/abs-2212-03657}, Question answering~\cite{DBLP:conf/ijcai/LiuH0018}, Reading Comprehension~\cite{DBLP:conf/acml/LiangLY19}. 
For CSC, although Self-Supervised Curriculum Learning has been employed in~\cite{DBLP:conf/emnlp/GanXZ21}, it is only integrated into a particular model. In our work, the universality of Curriculum Learning is the first time to be demonstrated for CSC. And our designed CL framework is model-agnostic for most existing CSC models. Besides, different from~\cite{DBLP:conf/emnlp/GanXZ21}, our work focuses more on contextual similarity.

\section{Proposed approach}
\label{sec3}

\subsection{Study Motivation}
\label{sec:motivation}
The core of our work is how to make the CSC models explicitly pay more attention to the context of Chinese characters. Therefore, we propose to use contextual similarity as the metric to measure the difficulty of samples in curriculum learning.

Based on detailed observation, we get the following two obvious facts:
(1) \emph{A sample is more difficult if it has more wrong characters.}
(2) \emph{A sample is more difficult if the wrong character it contains is more similar to the corresponding correct character.}
According to these two facts, we design a specific difficulty evaluation strategy and propose the curriculum learning framework for CSC. 
More specifically, as shown in Figure~\ref{Method_Figure}, our curriculum learning framework is divided into two parts: \textbf{Difficulty Evaluation} and \textbf{Curriculum Arrangement}, which will be described in Sections~\ref{sec:difficulty} and~\ref{sec:curriculum}.

\subsection{Difficulty Evaluation}
\label{sec:difficulty}
In Difficulty Evaluation, we assign a difficulty score to each training sample in the whole training set $S$. For each sample, we employ an encoder $E(.)$ (e.g., BERT or other CSC models), to transform the characters in the wrong sequence $s_i$ and correct sequence $t_i$ to the corresponding contextual representations $E(s_i)$ and $E(t_i)$.

After obtaining the contextual representations of the wrong/correct sentence, we use the representation corresponding to the wrong position to calculate the cosine similarity, and then the similarities corresponding to all positions with an error are summed up as the difficulty score of the sample:
\begin{equation}
\small
    d_i=\sum_{j\in W_i}\dfrac{E(s_i)_j\cdot E(t_i)_j}{||E(s_i)_j||\cdot ||E(t_i)_j||},
\end{equation}
where $d_i$ is the difficulty score of $i$-th sample, $W_i$ are the positions with error.
% To verify the effectiveness of our designed difficulty scoring function, we present the case study in Section~\ref{Case_Study}.

\subsection{Curriculum Arrangement}
\label{sec:curriculum}
In this section, we describe an \textbf{Annealing} method to arrange all the training samples $S$ into an ordered curriculum based on the difficulty scores that are introduced in Section~\ref{sec:difficulty}.

Firstly, we sort all training samples in ascending order of their difficulty scores and split them into $k$ subsets $\{S_1,S_2,...,S_k\}$. Note that these subsets are non-overlapping for preventing over-fitting and improving the generalization performance.
Then we arrange a learning curriculum which contains $k+1$ training stages. At the $i$-th stage ($i\le k$), we further split each of the $k$ subsets $\{S_1,S_2,...,S_k\}$ into $k$ parts by order of difficulty. For each subset $S_j$, we obtain $\{S_{j,1}, S_{j,2},..., S_{j,k}\}$ and use the $i$-th part $S_{j,i}$ for this $i$-th stage, thus the final training set $C_i=\{S_{1,i}, S_{2,i},..., S_{k,i}\}$ is employed for the $i$-th stage. It is worth mentioning that the training set $C_i$ will be shuffled for maintaining local stochastics within $i$-th stage.

For the former $k$ stages, the model is trained on the $C_i$ for one epoch one after another to lead the model learning from easy to difficult. At the last stage (i.e., the $k+1$-th stage ), the model is trained on the whole training set $S$ for fitting the original data distribution.

\section{Experiments}
\label{sec4}

\subsection{Datasets}
\label{data_prep}
Following previous works~\cite{DBLP:conf/acl/ZhangHLL20,DBLP:conf/acl/XuLZLWCHM21}, we use the same training data which contains SIGHAN 13/14/15~\cite{DBLP:conf/acl-sighan/WuLL13,yu2014chinese,DBLP:conf/acl-sighan/TsengLCC15} and a generated pseudo dataset~\cite{DBLP:conf/emnlp/WangSLHZ18}.
Additionally, to ensure fairness, models’ performance is evaluated on the same test data as our baselines, from the test datasets of SIGHAN 13/14/15.

\subsection{Baseline Methods}
We select strong confusion set-based models as baselines: \textbf{SpellGCN}~\cite{DBLP:conf/acl/ChengXCJWWCQ20} applies GCNs to learn the character similarity from confusion set. \textbf{PLOME}~\cite{DBLP:conf/acl/LiuYYZW20} designs pre-training strategy based on the confusion set. 
\textbf{MLM-phonetics}\cite{DBLP:conf/acl/ZhangPZWHSWW21} introduces phonetic similarity into masked language models from confusion set. 
\textbf{REALISE}~\cite{DBLP:conf/acl/XuLZLWCHM21} extracts and mixes semantic, phonetic, and graphic information.
In addition, we select three popular CSC models to be combined with our proposed CL method: \textbf{BERT}~\cite{DBLP:conf/naacl/DevlinCLT19} directly fine-tunes the $\text{BERT}_{\text{BASE}}$ on the CSC training data. \textbf{Soft-Masked BERT}~\cite{DBLP:conf/acl/ZhangHLL20} utilizes the confusion set to generate sufficient training data.  \textbf{MacBERT}~\cite{DBLP:conf/emnlp/CuiC000H20} improves the masking strategy of BERT and adds a full connection layer to detect errors. These BERT-based CSC models are all convenient to obtain the contextual representations of Chinese characters.

\subsection{Experimental Setup}
\label{sec:setup_appendix}
During the training phase, we follow the hyper-parameters of Soft-Masked-Bert ~\cite{DBLP:conf/acl/ZhangHLL20}. 
% For REALISE, the learning-rate is set to $5e-5$, the batch size is set to 32. In addition, warming up and linear decay are utilized during the model training with the AdamW optimizer.
For Soft-Masked-BERT
, we maintain a learning rate $2e-5$ and fine-tune the parameters with Adam. As for MacBERT~\cite{DBLP:conf/emnlp/CuiC000H20}, the learning-rate is set to $5e-5$, the batch size is set to 32. The $k$ is empirically set to in our method. During the testing phase, we evaluate the models in both detection and correction utilizing sentence-level accuracy, precision, recall, and F1 score. 

\subsection{Experimental Results}

Table~\ref{Main_Results} shows the results of our CL method compared to baselines. We can see that, by reordering the training data, our method yields consistent gain with a large margin against all baselines.

Particularly, unlike most models which leverage character similarity, our method achieves better performance by explicitly focusing on contextual similarity. Besides, the significant improvements over three models (i.e., Soft-Masked BERT, BERT, and MacBERT) verify the model-agnostic characteristic of our framework. 

\begin{table}[t]
    \centering
    \small
    \begin{tabular}{l|c|c}
        \toprule
          \textbf{Method} & Correction F1 & $\Delta$\\
         \midrule
         MacBERT & 75.0\% & -- \\
         Only Difficulty Evaluation & 77.2\% & +2.2\% \\
         Only Curriculum Arrangement & 76.3\% & +1.3\% \\
         Using Character Similarity & 76.5\% & +1.5\% \\
        \hline
         CL (MacBERT) & \textbf{78.4\%} & \textbf{+3.4\%} \\
         \bottomrule
    \end{tabular}
    \vspace{-0.15cm}
    \caption{Results of ablation studies. "$\Delta$" indicates the absolute F1 improvements on correction level.}
    \label{ablation_results}
    \vspace{-0.1cm}
\end{table}

\subsection{Ablation Study}
To explore the contribution of each component in our curriculum learning framework, we conduct ablation studies with the following settings: 1) \textbf{only using difficulty evaluation} to sort training samples for the training process of MacBERT, 2) \textbf{only using curriculum arrangement} to randomly arrange training stages for the training process of MacBERT, and the training samples of each stage are randomly selected. 3) Besides, to verify the advantage of contextual similarity, we also \textbf{use the confusion set-based character similarity} as the difficulty metric in our CL framework.

From Table~\ref{ablation_results}, we observe that both difficulty evaluation and curriculum arrangement bring improvements for MacBERT, which indicates the rationality of these two modules we design. Particularly, the greater improvements that only using difficulty evaluation brings to MacBERT than only using curriculum arrangement and using character similarity reflects the correctness of our motivation that contextual similarity is more valuable than character similarity in the CSC task.

\begin{figure}[t]
    \centering
    \includegraphics[width=0.50\textwidth]{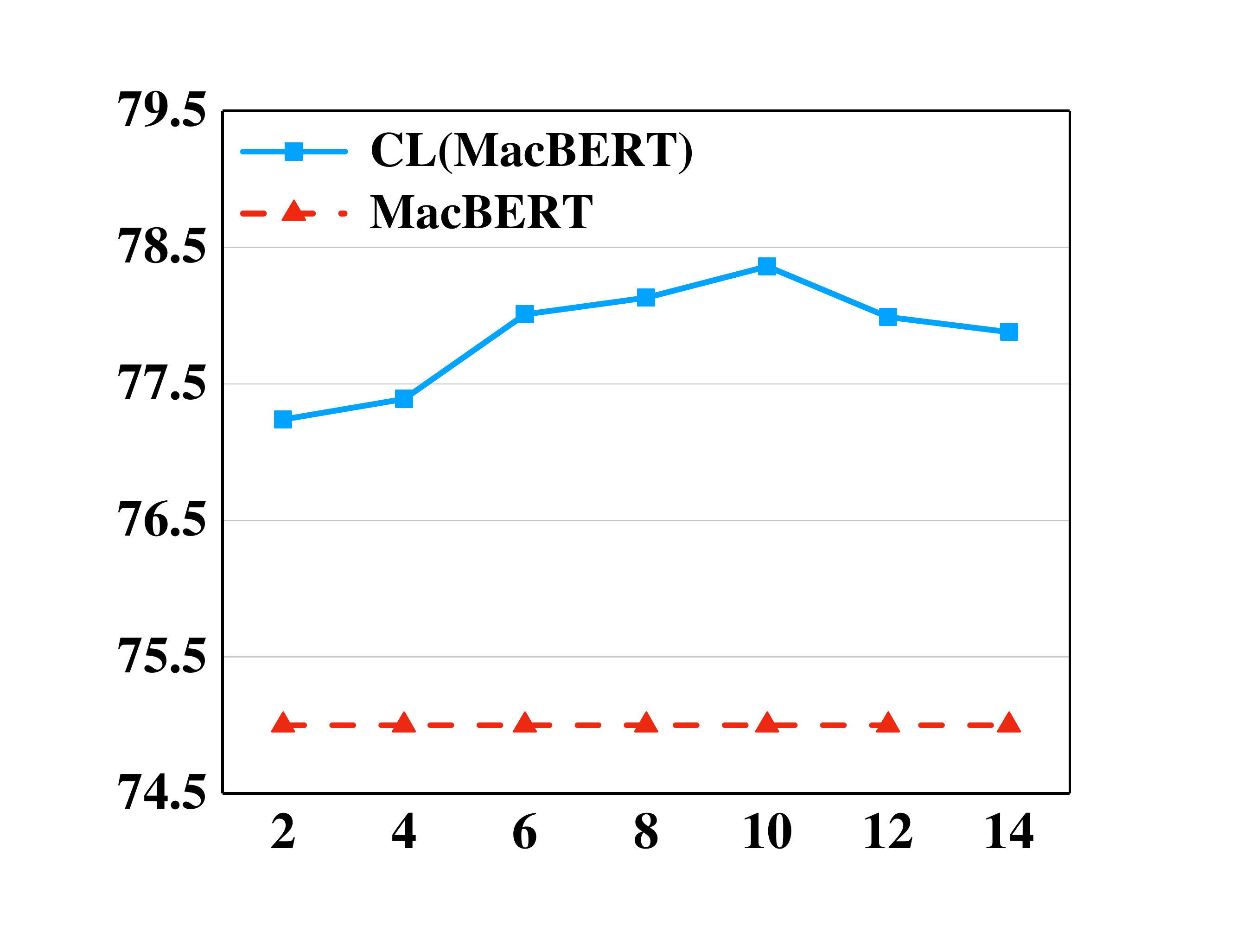}
    \caption{The correction F1 scores on SIGHAN15, using different values of $k$ in CL (MacBERT).}
    \label{different-buckets}
\end{figure}

\subsection{Parameter Study} 

The key parameter in our framework is the number of subsets $k$, so it is essential to study its effects. Figure~\ref{different-buckets} illustrates the performance change of CL~(MacBERT), we find that when the value of $k$ reaches a certain value, the performance of the model does not improve anymore. In fact, this phenomenon is consistent with the process of human learning. Imagine that when the courses are divided too trivially (that is, when the value of $k$ is too large), it is difficult for humans to learn effective knowledge from too many courses. Therefore, it is critical to choose the best $k$, although there are stable improvements based on MacBERT at all values of $k$.

\subsection{Case Study}
\label{Case_Study}
\begin{CJK*}{UTF8}{gbsn}
\begin{table}[t]
\small
\centering
\begin{tabular}{p{1.0cm}|p{4.0cm}|p{1.1cm}}
\toprule
  
% Case 2: \\ \midrule

\textbf{Input:} & 今天的夕阳真\textcolor{red}{每\textit{\pinyin{(mei3)}}}啊& \textbf{Difficulty}\\

\textbf{Trans:}&The sunset today is \textcolor{red}{common}&\\
\midrule
% \cmidrule(l){1-2} 
\textbf{Correct:} & 今天的夕阳真\textcolor{blue}{美\textit{\pinyin{(mei3)}}}啊&  \\
\textbf{Trans:} &The sunset today is \textcolor{blue}{beautiful}&0.612 \\
% 我要那个时候走/
\midrule
\textbf{Input:} & 你们可以走路或\textcolor{red}{造\textit{\pinyin{(zao4)}}}公交& \textbf{Difficulty}\\
% \textbf{Input:} & 他将\textcolor{red}{住\textit{\pinyin{(zhu4)}}}演这部作品。& \textbf{Diffulty}\\
\textbf{Trans:}&You can walk or \textcolor{red}{make} a bus&\\

\midrule

\textbf{Correct:} & 你们可以走路或\textcolor{blue}{坐\textit{\pinyin{(zuo4)}}}公交&\\

\textbf{Trans:}&You can walk or \textcolor{blue}{take} a bus&0.934\\

\bottomrule

\end{tabular}
\caption{Examples of spelling errors with low and high difficulty score. 
We mark the \textcolor{red}{input wrong}/\textcolor{blue}{correct} characters in \textcolor{red}{red}/\textcolor{blue}{blue}.  }
\label{Case_Studies}

\end{table}
In the first example of Table~\ref{Case_Studies}, “每(\textit{\pinyin{mei3}})” is the wrong character，“美(\textit{\pinyin{mei3}})” is the corrected character. The two characters are only  acoustically similar. Therefore, the low difficulty score evaluated by our CL(MacBERT) of this example is reasonable. 
In the second example, the erroneous character “造({\pinyin{zao4}})” is easily to be detected and corrected to "坐({\pinyin{zuo4}})" by using the information of "走路(walk)" as well as "公交(bus)" in the context.  
The difficulty score of the latter example is also significantly higher than the first example. 
This verifies the effectiveness of our designed difficulty scoring function and suggests that the model has learned how to distinguish between highly and slightly similar spelling errors.
\end{CJK*}

\section{Conclusion}
\label{sec5}

In this paper, we aim to exploit the contextual similarity of characters to obtain better CSC performance than character similarity contained in traditional confusion set. Additionally, we propose a simple yet effective curriculum learning framework for the CSC task. With the help of such a model-agnostic framework, most existing CSC models significantly perform better. In the future, it would be a very interesting direction to apply the core idea of our work to more scenarios such as Chinese grammatical error correction.

\section*{Acknowledgements}
This research is supported by the National Natural Science Foundation of China (Grant No.62276154), AMiner.Shenzhen SciBrain Fund, Research Center for Computer Network (Shenzhen) Ministry of Education, Beijing Academy of Artificial Intelligence (BAAI), the Natural Science Foundation of Guangdong Province (Grant No. 2021A1515012640), Basic Research Fund of Shenzhen City (Grant No. JCYJ20210324120012033 and JSGG20210802154402007), and Overseas Cooperation Research Fund of Tsinghua Shenzhen International Graduate School (Grant No. HW2021008).

\clearpage
\bibliography{anthology,custom}

\begin{thebibliography}{27}
\expandafter\ifx\csname natexlab\endcsname\relax\def\natexlab#1{#1}\fi

\bibitem[{Afli et~al.(2016)Afli, Qiu, Way, and
  Sheridan}]{DBLP:conf/lrec/AfliQWS16}
Haithem Afli, Zhengwei Qiu, Andy Way, and P{\'{a}}raic Sheridan. 2016.
\newblock Using {SMT} for {OCR} error correction of historical texts.
\newblock In \emph{{LREC}}. European Language Resources Association {(ELRA)}.

\bibitem[{Bengio et~al.(2009)Bengio, Louradour, Collobert, and
  Weston}]{DBLP:conf/icml/BengioLCW09}
Yoshua Bengio, J{\'{e}}r{\^{o}}me Louradour, Ronan Collobert, and Jason Weston.
  2009.
\newblock Curriculum learning.
\newblock In \emph{{ICML}}, volume 382 of \emph{{ACM} International Conference
  Proceeding Series}, pages 41--48. {ACM}.

\bibitem[{Cheng et~al.(2020)Cheng, Xu, Chen, Jiang, Wang, Wang, Chu, and
  Qi}]{DBLP:conf/acl/ChengXCJWWCQ20}
Xingyi Cheng, Weidi Xu, Kunlong Chen, Shaohua Jiang, Feng Wang, Taifeng Wang,
  Wei Chu, and Yuan Qi. 2020.
\newblock Spellgcn: Incorporating phonological and visual similarities into
  language models for chinese spelling check.
\newblock In \emph{{ACL}}, pages 871--881. Association for Computational
  Linguistics.

\bibitem[{Cheng et~al.(2022)Cheng, Dong, Yue, Ko, Wang, and
  Zou}]{DBLP:journals/corr/abs-2212-03657}
Xuxin Cheng, Qianqian Dong, Fengpeng Yue, Tom Ko, Mingxuan Wang, and Yuexian
  Zou. 2022.
\newblock \href {https://doi.org/10.48550/arXiv.2212.03657} {{M3ST:} mix at
  three levels for speech translation}.
\newblock \emph{CoRR}, abs/2212.03657.

\bibitem[{Cheng et~al.(2023)Cheng, Zhu, Li, Li, and
  Zou}]{DBLP:journals/corr/abs-2302-09328}
Xuxin Cheng, Zhihong Zhu, Hongxiang Li, Yaowei Li, and Yuexian Zou. 2023.
\newblock \href {https://doi.org/10.48550/arXiv.2302.09328} {{SSVMR:}
  saliency-based self-training for video-music retrieval}.
\newblock \emph{CoRR}, abs/2302.09328.

\bibitem[{Cui et~al.(2020)Cui, Che, Liu, Qin, Wang, and
  Hu}]{DBLP:conf/emnlp/CuiC000H20}
Yiming Cui, Wanxiang Che, Ting Liu, Bing Qin, Shijin Wang, and Guoping Hu.
  2020.
\newblock Revisiting pre-trained models for chinese natural language
  processing.
\newblock In \emph{{EMNLP} (Findings)}, volume {EMNLP} 2020 of \emph{Findings
  of {ACL}}, pages 657--668. Association for Computational Linguistics.

\bibitem[{Devlin et~al.(2019)Devlin, Chang, Lee, and
  Toutanova}]{DBLP:conf/naacl/DevlinCLT19}
Jacob Devlin, Ming{-}Wei Chang, Kenton Lee, and Kristina Toutanova. 2019.
\newblock {BERT:} pre-training of deep bidirectional transformers for language
  understanding.
\newblock In \emph{{NAACL-HLT} {(1)}}, pages 4171--4186. Association for
  Computational Linguistics.

\bibitem[{Dong and Zhang(2016)}]{DBLP:conf/emnlp/DongZ16}
Fei Dong and Yue Zhang. 2016.
\newblock Automatic features for essay scoring - an empirical study.
\newblock In \emph{{EMNLP}}, pages 1072--1077. The Association for
  Computational Linguistics.

\bibitem[{Gan et~al.(2021)Gan, Xu, and Zan}]{DBLP:conf/emnlp/GanXZ21}
Zifa Gan, Hongfei Xu, and Hongying Zan. 2021.
\newblock Self-supervised curriculum learning for spelling error correction.
\newblock In \emph{{EMNLP} {(1)}}, pages 3487--3494. Association for
  Computational Linguistics.

\bibitem[{Huang et~al.(2021)Huang, Li, Jiang, Zhang, Chen, Wang, and
  Xiao}]{DBLP:conf/acl/HuangLJZCWX20}
Li~Huang, Junjie Li, Weiwei Jiang, Zhiyu Zhang, Minchuan Chen, Shaojun Wang,
  and Jing Xiao. 2021.
\newblock Phmospell: Phonological and morphological knowledge guided chinese
  spelling check.
\newblock In \emph{{ACL/IJCNLP} {(1)}}, pages 5958--5967. Association for
  Computational Linguistics.

\bibitem[{Kocmi and Bojar(2017)}]{kocmi2017curriculum}
Tom Kocmi and Ondrej Bojar. 2017.
\newblock Curriculum learning and minibatch bucketing in neural machine
  translation.
\newblock \emph{arXiv preprint arXiv:1707.09533}.

\bibitem[{Li et~al.(2023)Li, Cao, Cheng, Zhu, Li, and
  Zou}]{DBLP:journals/corr/abs-2301-05997}
Hongxiang Li, Meng Cao, Xuxin Cheng, Zhihong Zhu, Yaowei Li, and Yuexian Zou.
  2023.
\newblock \href {https://doi.org/10.48550/arXiv.2301.05997} {Generating
  templated caption for video grounding}.
\newblock \emph{CoRR}, abs/2301.05997.

\bibitem[{Li et~al.(2022{\natexlab{a}})Li, Ma, Zhou, Li, Yangning, Huang, Liu,
  Li, Cao, and Zheng}]{li-etal-2022-learning-dictionary}
Yinghui Li, Shirong Ma, Qingyu Zhou, Zhongli Li, Li~Yangning, Shulin Huang,
  Ruiyang Liu, Chao Li, Yunbo Cao, and Haitao Zheng. 2022{\natexlab{a}}.
\newblock \href {https://aclanthology.org/2022.findings-emnlp.18} {Learning
  from the dictionary: Heterogeneous knowledge guided fine-tuning for {C}hinese
  spell checking}.
\newblock In \emph{Findings of the Association for Computational Linguistics:
  EMNLP 2022}, pages 238--249, Abu Dhabi, United Arab Emirates. Association for
  Computational Linguistics.

\bibitem[{Li et~al.(2022{\natexlab{b}})Li, Zhou, Li, Li, Liu, Sun, Wang, Li,
  Cao, and Zheng}]{DBLP:conf/acl/LiZLLLSWLCZ22}
Yinghui Li, Qingyu Zhou, Yangning Li, Zhongli Li, Ruiyang Liu, Rongyi Sun,
  Zizhen Wang, Chao Li, Yunbo Cao, and Hai{-}Tao Zheng. 2022{\natexlab{b}}.
\newblock \href {https://doi.org/10.18653/v1/2022.findings-acl.252} {The past
  mistake is the future wisdom: Error-driven contrastive probability
  optimization for chinese spell checking}.
\newblock In \emph{Findings of the Association for Computational Linguistics:
  {ACL} 2022, Dublin, Ireland, May 22-27, 2022}, pages 3202--3213. Association
  for Computational Linguistics.

\bibitem[{Liang et~al.(2019)Liang, Li, and Yin}]{DBLP:conf/acml/LiangLY19}
Yichan Liang, Jianheng Li, and Jian Yin. 2019.
\newblock A new multi-choice reading comprehension dataset for curriculum
  learning.
\newblock In \emph{{ACML}}, volume 101 of \emph{Proceedings of Machine Learning
  Research}, pages 742--757. {PMLR}.

\bibitem[{Liu et~al.(2018)Liu, He, Liu, and Zhao}]{DBLP:conf/ijcai/LiuH0018}
Cao Liu, Shizhu He, Kang Liu, and Jun Zhao. 2018.
\newblock Curriculum learning for natural answer generation.
\newblock In \emph{{IJCAI}}, pages 4223--4229. ijcai.org.

\bibitem[{Liu et~al.(2021)Liu, Yang, Yue, Zhang, and
  Wang}]{DBLP:conf/acl/LiuYYZW20}
Shulin Liu, Tao Yang, Tianchi Yue, Feng Zhang, and Di~Wang. 2021.
\newblock {PLOME:} pre-training with misspelled knowledge for chinese spelling
  correction.
\newblock In \emph{{ACL/IJCNLP} {(1)}}, pages 2991--3000. Association for
  Computational Linguistics.

\bibitem[{Ma et~al.(2022)Ma, Li, Sun, Zhou, Huang, Zhang, Yangning, Liu, Li,
  Cao, Zheng, and Shen}]{ma-etal-2022-linguistic}
Shirong Ma, Yinghui Li, Rongyi Sun, Qingyu Zhou, Shulin Huang, Ding Zhang,
  Li~Yangning, Ruiyang Liu, Zhongli Li, Yunbo Cao, Haitao Zheng, and Ying Shen.
  2022.
\newblock \href {https://aclanthology.org/2022.findings-emnlp.40} {Linguistic
  rules-based corpus generation for native {C}hinese grammatical error
  correction}.
\newblock In \emph{Findings of the Association for Computational Linguistics:
  EMNLP 2022}, pages 576--589, Abu Dhabi, United Arab Emirates. Association for
  Computational Linguistics.

\bibitem[{Soviany et~al.(2021)Soviany, Ionescu, Rota, and
  Sebe}]{soviany2021curriculum}
Petru Soviany, Radu~Tudor Ionescu, Paolo Rota, and Nicu Sebe. 2021.
\newblock Curriculum learning: A survey.
\newblock \emph{arXiv preprint arXiv:2101.10382}.

\bibitem[{Tseng et~al.(2015)Tseng, Lee, Chang, and
  Chen}]{DBLP:conf/acl-sighan/TsengLCC15}
Yuen{-}Hsien Tseng, Lung{-}Hao Lee, Li{-}Ping Chang, and Hsin{-}Hsi Chen. 2015.
\newblock Introduction to {SIGHAN} 2015 bake-off for chinese spelling check.
\newblock In \emph{SIGHAN@IJCNLP}, pages 32--37. Association for Computational
  Linguistics.

\bibitem[{Wang et~al.(2018)Wang, Song, Li, Han, and
  Zhang}]{DBLP:conf/emnlp/WangSLHZ18}
Dingmin Wang, Yan Song, Jing Li, Jialong Han, and Haisong Zhang. 2018.
\newblock A hybrid approach to automatic corpus generation for chinese spelling
  check.
\newblock In \emph{{EMNLP}}, pages 2517--2527. Association for Computational
  Linguistics.

\bibitem[{Wu et~al.(2013)Wu, Liu, and Lee}]{DBLP:conf/acl-sighan/WuLL13}
Shih{-}Hung Wu, Chao{-}Lin Liu, and Lung{-}Hao Lee. 2013.
\newblock Chinese spelling check evaluation at {SIGHAN} bake-off 2013.
\newblock In \emph{SIGHAN@IJCNLP}, pages 35--42. Asian Federation of Natural
  Language Processing.

\bibitem[{Xu et~al.(2021)Xu, Li, Zhou, Li, Wang, Cao, Huang, and
  Mao}]{DBLP:conf/acl/XuLZLWCHM21}
Heng{-}Da Xu, Zhongli Li, Qingyu Zhou, Chao Li, Zizhen Wang, Yunbo Cao, Heyan
  Huang, and Xian{-}Ling Mao. 2021.
\newblock Read, listen, and see: Leveraging multimodal information helps
  chinese spell checking.
\newblock In \emph{{ACL/IJCNLP} (Findings)}, volume {ACL/IJCNLP} 2021 of
  \emph{Findings of {ACL}}, pages 716--728. Association for Computational
  Linguistics.

\bibitem[{Yu and Li(2014)}]{yu2014chinese}
Junjie Yu and Zhenghua Li. 2014.
\newblock Chinese spelling error detection and correction based on language
  model, pronunciation, and shape.
\newblock \emph{CLP 2014}, page 220.

\bibitem[{Zhang et~al.(2021)Zhang, Pang, Zhang, Wang, He, Sun, Wu, and
  Wang}]{DBLP:conf/acl/ZhangPZWHSWW21}
Ruiqing Zhang, Chao Pang, Chuanqiang Zhang, Shuohuan Wang, Zhongjun He, Yu~Sun,
  Hua Wu, and Haifeng Wang. 2021.
\newblock Correcting chinese spelling errors with phonetic pre-training.
\newblock In \emph{{ACL/IJCNLP} (Findings)}, volume {ACL/IJCNLP} 2021 of
  \emph{Findings of {ACL}}, pages 2250--2261. Association for Computational
  Linguistics.

\bibitem[{Zhang et~al.(2020)Zhang, Huang, Liu, and
  Li}]{DBLP:conf/acl/ZhangHLL20}
Shaohua Zhang, Haoran Huang, Jicong Liu, and Hang Li. 2020.
\newblock Spelling error correction with soft-masked {BERT}.
\newblock In \emph{{ACL}}, pages 882--890. Association for Computational
  Linguistics.

\bibitem[{Zhu et~al.(2022)Zhu, Xu, Cheng, Song, and
  Zou}]{DBLP:journals/corr/abs-2211-04023}
Zhihong Zhu, Weiyuan Xu, Xuxin Cheng, Tengtao Song, and Yuexian Zou. 2022.
\newblock \href {https://doi.org/10.48550/arXiv.2211.04023} {A dynamic graph
  interactive framework with label-semantic injection for spoken language
  understanding}.
\newblock \emph{CoRR}, abs/2211.04023.

\end{thebibliography}

\end{document}